\documentclass[11pt,a4paper]{article}
\usepackage[hyperref]{eacl2021}
\usepackage{times}
\usepackage{latexsym}

\usepackage{microtype}

\usepackage[ruled,vlined]{algorithm2e}
\usepackage{CJK}
\usepackage{multirow}
\usepackage{booktabs}
\usepackage{graphicx}
\usepackage{amsmath}
\usepackage{ulem}
\usepackage{balance}
\usepackage[colorinlistoftodos]{todonotes}

\aclfinalcopy 

\title{Quality Estimation without Human-labeled Data}

\author{Yi-Lin Tuan\textsuperscript{1}, Ahmed El-Kishky\textsuperscript{2}, Adithya Renduchintala\textsuperscript{2}, Vishrav Chaudhary\textsuperscript{2},\\ \textbf{Francisco Guzm\'an\textsuperscript{2} \and Lucia Specia\textsuperscript{3}} \\
  \textsuperscript{1}University of California Santa Barbara,
  \textsuperscript{2}Facebook AI,
  \textsuperscript{3}Imperial College London \\
  \texttt{\textsuperscript{1}ytuan@cs.ucsb.edu, \textsuperscript{2}\{ahelk,adirendu,vishrav,fguzman\}@fb.com}\\
  \texttt{\textsuperscript{3}l.specia@imperial.ac.uk}
  
  }

\date{}

\begin{document}
\maketitle

\begin{abstract}
Quality estimation aims to measure the quality of translated content without access to a reference translation. This is crucial for machine translation systems in real-world scenarios where high-quality translation is needed. While many approaches exist for quality estimation, they are based on supervised machine learning requiring costly human labelled data. As an alternative, we propose a technique that does not rely on examples from human-annotators and instead uses synthetic training data. We train off-the-shelf architectures for supervised quality estimation on our synthetic data and show that the resulting models achieve comparable performance to models trained on human-annotated data, both for sentence and word-level prediction.
\end{abstract}

\section{Introduction}
The adoption of Machine Translation (MT) has been increasing in areas ranging from government and 
finance, 
to even social media due to the substantial improvements achieved from Neural Machine Translation (NMT). However, even with improved performance, translation quality is not consistent across language pairs, domains, and sentences. This can be detrimental to end-user's trust and can cause unintended consequences arising from poor translations. 
Thus, having metrics to assess the quality of translated content is crucial to ensure that only high-quality translations are provided to end-users or 
downstream tasks. Quality Estimation (QE) metrics aim to predict translation quality without access to reference translations~\cite{blatz2004confidence,specia2009estimating, specia2013quest}.

State-of-the-art QE techniques 
have leveraged MT systems and language-specific human annotations as supervision, including direct assessment and post-editing \cite{kepler-etal-2019-unbabels,fonseca-etal-2019-findings,sun2020exploratory}. However, these annotations are costly and time-consuming, particularly for word-level QE, where each token needs a label. 

Some unsupervised approaches take inspiration from statistical MT~\cite{popovic-2012-morpheme, moreau2012quality, etchegoyhen2018supervised} or apply uncertainty quantification~\cite{fomicheva2020unsupervised} for QE. However, their performance is inferior to that of supervised models. In related areas such as automatic post-editing, parallel data has been used to create synthetic post-editing data~\cite{negri2018escape}, however this technique only compares machine-translated sentences to references. Our approach augments MT errors with additional errors via masked language model rewriting. 


We leverage noisy, mined comparable sentences obtained by weakly-supervised techniques~\cite{el2020searching}. These noisy bitexts have been mined from a variety of domains such as Wikipedia~\cite{schwenk2019wikimatrix} and large web-crawls~\cite{schwenk2019ccmatrix,elkishkyccaligned,el2020massively} and have been shown to be an invaluable source of training data for NMT models. Using this data is crucial to avoid data leakage between a trained NMT model and the data we use to create synthetic QE data. For each source-target sentence pair from the mined data, we apply an MT system to generate a candidate translation of the source sentence. Additionally we rewrite each target reference sentence using a masked language model to introduce errors. These two approaches generate two alternative ``translations'' of the source sentence. We then produce pseudo-labels for each token in these translations by edit distance alignment to the original reference sentence. 
This results in each translated word being pseudo-labelled as correct or incorrect, which is our synthetic 
QE training data. Analogously, sentence-level training data is derived as the proportion of incorrect words  per sentence.

Our {\bf main contributions} are: (i) We explore a simple technique to effectively generate synthetic data for QE that allows for both word-level and sentence-level estimation (ii) we demonstrate that our technique performs comparably to off-the-shelf models trained on human-annotated data.

\section{QE Task Description}

Word-level QE has been mainly framed as the task of predicting which words in the translation need to be post-edited. As such, word-level QE aims to assign a tag for each word and gap between words in a machine-generated translation as {\em correct}, i.e.,  the word does not need editing, or {\em incorrect}, i.e., the words should be substituted, deleted, or inserted (tags for gaps) \cite{specia-EtAl:2020:WMT2}. 

For word-level, we denote the tag of each word in a translation as ${m}_t\in \{{\tt OK}, {\tt BAD}\}$, where $t\in [1,T]$ and $T$ is the length of the translation.
Also, we denote the tag of each gap between two words (including the beginning and the end) as ${g}_t\in \{{\tt OK}, {\tt BAD}\}$, where $t\in [1, T+1]$.
\begin{figure}[t!]
\centering
\includegraphics[width=.97\linewidth]{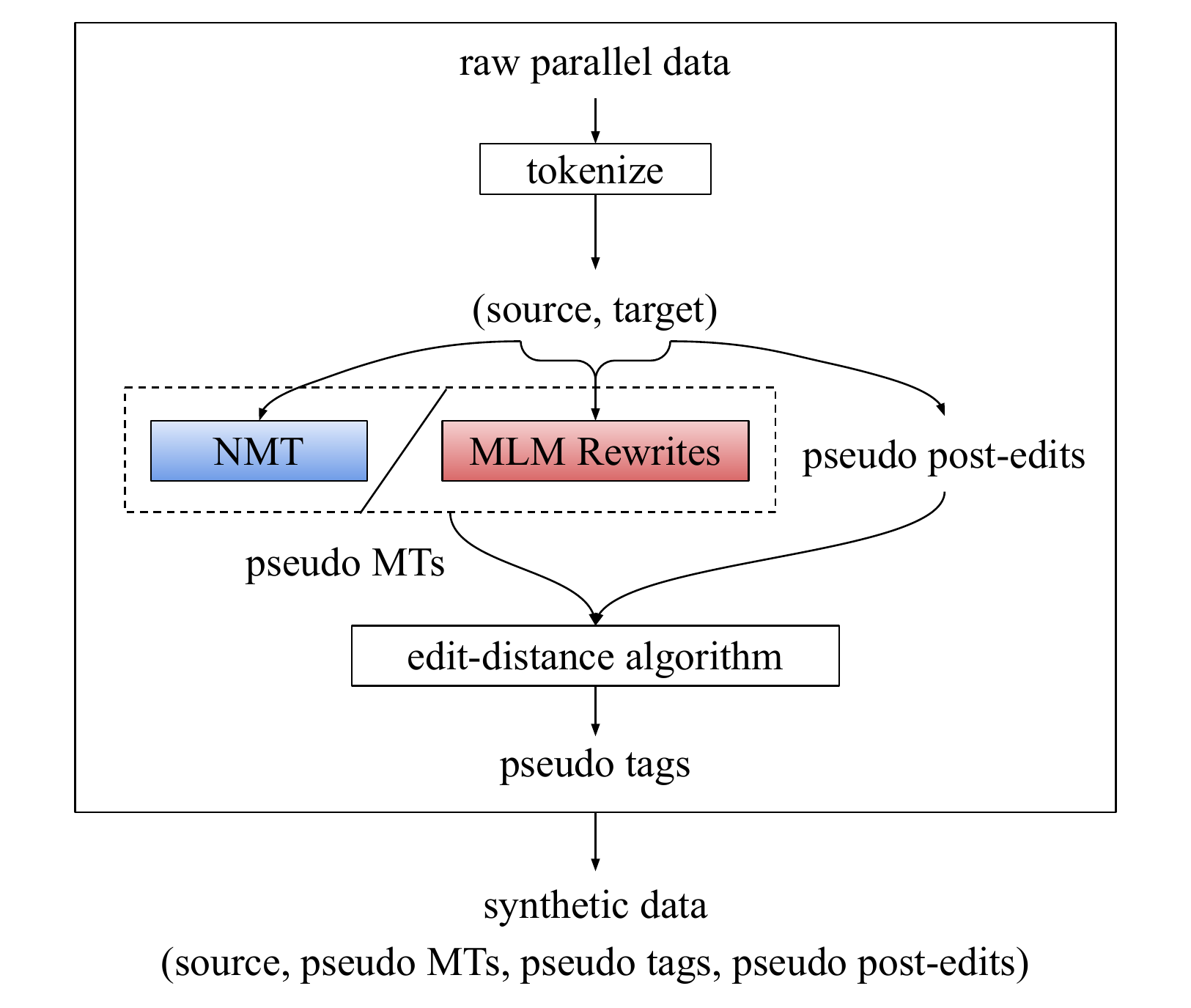}
\caption{The pipeline to synthesize data for QE from comparable mined data.}
\label{fig:pipeline}
\end{figure}

In traditional QE, data is collected by first translating source sentences using an MT model. Second, experts post-edit these translations. Third, the post-edits and machine translations are aligned in such a way that induces the minimum edit distance between the tokens of each.
Finally, each $m_t$ is labelled as {\tt BAD} if it should be deleted or substituted and each $g_t$ is labelled as {\tt BAD}  if at least a word should be inserted there. Sentence-level QE labels can be generated 
by computing the Human-targeted Translation Error Rate (HTER) \cite{snover-brent-2001-bayesian, Snover06astudy}, which is the minimum ratio of edit operations needed to fix the translation to the number of its tokens. We explore the possibility to skip the costly human post-editing process by proposing a data synthesis pipeline, which we then test on human labelled data.

\begin{figure}[t!]
    \centering
    \includegraphics[width=.97\linewidth]{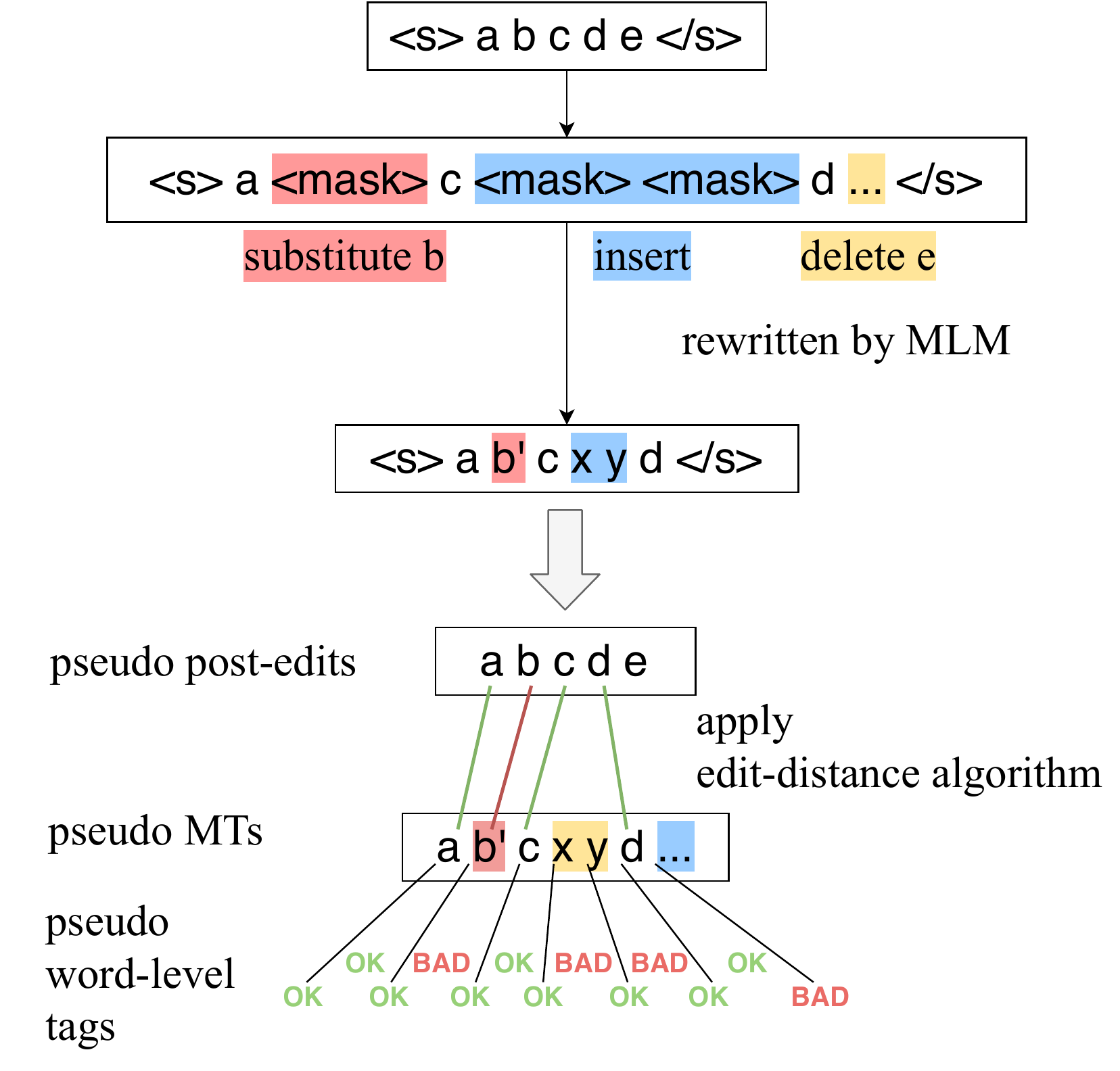}
    \caption{The rewriting process by text-infilling using a masked language model.}
    \label{fig:rewrites}
\end{figure}


\begin{table*}\small
    \centering
    \begin{tabular}{@{}lrccrcc@{}}
    \toprule
        \multirow{2}{*}{\bf Data} & \multicolumn{3}{c}{\bf English-German} & \multicolumn{3}{c}{\bf English-Chinese} 
        \\
        \cmidrule(r){2-4}
        \cmidrule(l){5-7}
        & size & 
        \begin{tabular}{@{}c@{}}MT bad (\%)\end{tabular}&  
        \begin{tabular}{@{}c@{}}Gap bad (\%)\end{tabular} & 
        size &
       \begin{tabular}{@{}c@{}}MT bad (\%)\end{tabular}&  
       \begin{tabular}{@{}c@{}}Gap bad (\%)\end{tabular} \\\midrule
        Human annotation &  7K  & 27.8 &4.7 & 7K &54.2  &8.4 \\
       NMT & 459K &38.2  &5.7 & 189K& 49.5 & 6.8 \\ 
        MLM (word-QE) & 459K &40.7 & 2.9& 189K& 53.9&8.6 \\
       MLM (sent-QE) & 459K & 43.1 & 3.3& 189K& 49.9& 2.7\\
        
        \bottomrule[1pt]
    \end{tabular}
    \caption{Statistics of annotated and synthetic (NMT and MLM) data.}
    \label{tab:data}
\vspace{-10pt}
\end{table*}

\section{Approach to Data Synthesis}
As depicted in Figure~\ref{fig:pipeline}, we synthesize data from mined Wikipedia datasets, where each example consists of a \emph{(source, target)} sentence pair.

We create candidate translations of source sentences in two ways:
For the first approach, we apply the \emph{NMT} model to translate each source sentence. For the second approach, we rewrite each reference target sentence using a masked language model (MLM), as shown in the \emph{MLM Rewrites} block in Figure~\ref{fig:pipeline}.
The two approaches create two forms of translations. Then, by treating target sentences as if they were post-edited data (\emph{pseudo post-edits}), we identify errors in each candidate translation by looking at the insertions, deletions, and substitutions 
between the references and generated translations.

\paragraph{Neural Machine Translation.}
For the first approach to generating synthetic data, we use a pretrained NMT model to create translations. The NMT model is the same model that was used to generate translations in the supervised data; the architecture is a standard transformer as used in~\cite{vaswani2017attention,ott2019fairseq}. The process of creating synthetic QE data first involves translating each source sentence using this model and taking the output as a translation which will later be used to generate the synthetic labels. When decoding, we apply a beam of 5 following the NMT models available in~\citet{fomicheva2020unsupervised} to generate a candidate translation. Next, we take the mined reference target sentence and treat it as a pseudo post-editing of the machine translation. 

We then compute the edit distance between MTs and pseudo post-edits. The resulted edit operations are the pseudo tags, which consist of word tags $m_t$ and gap tags $g_t$.
This process is illustrated in Algorithm~\ref{alg:nmt}.

\begin{algorithm}\small
  \caption{DataSynthesis-NMT}
  \label{alg:nmt}
  \KwIn{pairs (source, target) from mined data, pretrained NMT model}
  \KwOut{(MTs, pseudo tags)}
  \For{each pair (source, target)}
  {
    MTs = NMT(source)\\
    $\{m_t\}_{t=1}^T$, $\{g_t\}_{t=1}^{T+1}$ = edit\_distance(MTs, target)\\
    pseudo tags = ($\{m_t\}_{t=1}^T$, $\{g_t\}_{t=1}^{T+1}$)\\
    return (MTs, pseudo tags)\\
  }

\end{algorithm}

\paragraph{Rewriting by Masked Language Model (MLM).}
Our second approach to creating synthetic QE training data is to introduce errors by rewriting target sentences. We inject these errors by performing \textit{text-infilling}~\cite{zhu2019text, lewis2019bart}. As displayed in Figure~\ref{fig:rewrites}, we perform text-infilling by applying three operations: (1) randomly substituting a proportion of tokens with a {\tt <mask>} token, (2) deleting consecutive tokens, and (3) inserting additional consecutive {\tt <mask>} tokens. We determine the lengths of consecutive deletions and insertions by drawing them from a Poisson distribution with mean $\lambda=1$ shifted by 1 to avoid zero-length insertions or deletions. We then use a pre-trained masked language model (MLM) supplied with the source sentence as input to infill the masked reference sentence. We select multilingual BERT~\cite{devlin2019bert} as it is pre-trained on Wikipedia which is in-domain to our test set. We present the target-rewriting approach in detail in Algorithm~\ref{alg:mlm}.

\begin{algorithm}\small
  \caption{DataSynthesis-Rewriting}
  \label{alg:mlm}
  \KwIn{pairs $(S,W)$:=(source, target) from mined data, pretrained MLM}
  \KwIn{$P_s$, $P_d$, $P_i$ as the probabilities of substitution, deletion, and insertion}
  \KwOut{(pseudo MTs, pseudo tags)}
  \For{each pair $(S,W)$}
  {
    $W'$ = randomly mask tokens in $W$ by $P_s$\\
    $D$ = randomly mark deletion in $W'$ by $P_d$\\
    $W'$ = randomly delete a text span from marks $D$ in $W'$ (length$\sim Poisson(\lambda=1)+1$)\\
    $I$ = randomly mark insertion in $W'$ by $P_i$\\
    $W'$ = randomly insert contiguous masks from marks $I$ in $W'$ (length$\sim Poisson(\lambda=1)+1$)\\
    rewrites = MLM\_fills\_in\_masks($S, W'$)\\
    $\{m_t\}_{t=1}^T$, $\{g_t\}_{t=1}^T$ = edit\_distance(rewrites, $W$)\\
    pseudo tags = ($\{m_t\}_{t=1}^T$, $\{g_t\}_{t=1}^{T+1}$)\\
    return (rewrites, pseudo tags)\\
  }
\end{algorithm}

In Section~\ref{sec:experiments}, we will investigate the performance of QE models trained on NMT-based synthetic data, rewriter-based synthetic data, and a two-model ensemble where each model is trained on a different form of synthetic data.

\begin{table*}\small
\centering
\begin{tabular}{@{}lcccccc@{}}
\toprule
\multicolumn{1}{c}{\multirow{2}{*}{\bf Data }} & \multicolumn{3}{c}{\bf English-German} & \multicolumn{3}{c}{\bf English-Chinese} \\
\cmidrule(r){2-4}
\cmidrule(l){5-7}
\multicolumn{1}{c}{} & MCC & F1-Ok & F1-Bad & MCC & F1-Ok & F1-Bad \\ \midrule 
Human annotation & \textbf{0.399} & 0.879 & \textbf{0.495} & 0.525 & 0.820 & 0.659 \\
MLM & 0.332 & \textbf{0.892} & 0.438 & 0.500 & 0.850 & 0.643 \\
NMT & 0.379 & 0.826 & 0.468 & 0.525 & \textbf{0.859} & 0.660 \\
NMT + MLM & \textbf{0.399} & 0.866 & 0.493 & \textbf{0.546} & 0.835 & \textbf{0.675}\\
\hline
Improvement (\%) & +0.20 & -1.40 & -0.40 & +4.00 & +1.83 & +2.43\\
\bottomrule
\end{tabular}
\caption{Results of word-level QE trained on human-annotated (7k) and synthetic data. Improvement in MCC for en-de \& en-zh shows synthetic data can train word-level models comparable to human-annotated data. We report improvement comparing models trained with human-annotation vs our combined NMT+MLM synthetic data.}
\label{tab:main-test}
\end{table*}

\begin{table*}\small
\centering
\begin{tabular}{@{}lllllll@{}}
\toprule
\multicolumn{1}{c}{\multirow{2}{*}{\bf Data}} & \multicolumn{3}{c}{\bf English-German} & \multicolumn{3}{c}{\bf English-Chinese} \\
\cmidrule(r){2-4}
\cmidrule(l){5-7}
\multicolumn{1}{c}{} & Pearson & MAE & RMSE & Pearson & MAE & RMSE\\ \midrule
Human annotation & \textbf{0.394} & \textbf{0.150} & \textbf{0.187} & 0.490 & 0.151 & 0.186 \\
MLM & 0.290 & 0.156 & 0.195 & 0.418 & 0.224 & 0.269 \\
NMT & 0.327 & 0.229 & 0.270 & 0.482 & 0.161 & 0.203\\
NMT + MLM & 0.373 & 0.172 & 0.205 & \textbf{0.506} & \textbf{0.148} & \textbf{0.183}\\
\hline
Improvement (\%) & -5.50 & +14.7 & +9.63 & +3.18 & -1.79 & -1.67\\
\bottomrule
\end{tabular}
\caption{Results of sentence-level HTER QE trained on human-annotated and synthetic data. For Pearson, positive improvement is better while for MAE \& RMSE negative is better. We report improvement comparing models trained with human-annotation vs our combined NMT+MLM synthetic data.}
\label{tab:main-sent-test}
\end{table*}

\section{Experiments and Results}\label{sec:experiments}
We focus on data released by the WMT20 shared task on QE for predicting post-editing effort, which includes English-to-German (En-De) and English-to-Chinese (En-Zh) word-level data and their sentence-level HTER \cite{specia-EtAl:2020:WMT2}.\footnote{Available here: https://github.com/sheffieldnlp/mlqe-pe} As the human-annotated data is sampled from Wikipedia, we choose to synthesize data from WikiMatrix~\cite{schwenk2019wikimatrix}, which consists of mined Wikipedia parallel data from which we sample pairs with a LASER~\cite{artetxe2019massively} margin score threshold of $1.06$ to ensure high-quality pairs. We note that the original QE data is not a subset of WikiMatrix. The German and Chinese text were tokenized using the Moses\footnote{https://github.com/alvations/sacremoses} and Jieba\footnote{https://github.com/fxsjy/jieba} tokenizers, respectively.
We list the statistics of the filtered Wikimatrix data as well as our resulting synthetic data in Table~\ref{tab:data}.

For the off-the-shelf QE model, we choose the multi-task predictor-estimator model~\cite{kim2017predictor} implemented by OpenKiwi v0.1.3~\cite{kepler2019openkiwi}.
This was the top-performing architecture for QE at WMT19 \cite{kepler-etal-2019-unbabels,fonseca-etal-2019-findings}. We train the predictor on parallel MT data provided by the WMT20 QE shared task. The predictor reads in words' contextualized word representations, the estimator passes these features through a 2-layer 125-dimension bidirectional LSTM (biLSTM) and then feeds the outputs into 1-layer linear word-level classifier. The first output of the biLSTM is also fed into a multi-layer perceptron to predict a sentence-level score. For multi-task learning, we train the model with both word- and sentence-level data.

For a fair comparison, we take the pre-trained predictor provided by the WMT20 QE shared task, fine-tune the whole model on the human annotated data, and compare results to those when fine-tuned on our synthetic data. We test by comparing model predictions and held-out human-annotated QE at word and sentence-level. At the word level, we measure QE performance with Matthew's Correlation Coefficient (MCC)~\cite{matthews1975comparison} (main metric), as well as F1 scores for BAD and OK tags. At the sentence-level, we measure the sentence-level Pearson's correlation~\cite{benesty2009pearson}, mean absolute error (MAE) and Root-mean-square deviation (RMSE). 

As shown in Table~\ref{tab:main-test}, for word-level QE,\footnote{The results reported in Tables~\ref{tab:main-test} and \ref{tab:main-sent-test} are evaluated on the test set provided (test20).} the model trained on synthetic data generated from NMT translations performs comparably to the same model trained on the original 7k human-annotated post-edits. This suggests that having human annotators post-edit each translation to create training data may be unnecessary and using reference sentences is good enough. The model trained on the MLM rewriting synthetic data generally under-performs compared to NMT generated data on MCC. However, we note that it performs better on F1 on OK tags. Therefore, we also ensemble the two models trained on each set of synthetic data through a linear combination. This yields comparable or better performance than the model trained on human-annotated data according to the main metric, MCC. 

In Table~\ref{tab:main-sent-test}, we compare the models trained on human-annotated data to our synthetic data for predicting sentence-level HTER scores. Again our synthetic data from NMT-generated translations outperforms MLM-rewriting data. Both under-perform  models trained on human-annotated data, but when combined they significantly improve and even outperform human-annotated for En-Zh. This once again suggests that the two forms of synthetic data are complementary and provide valuable signals for QE.

\section {Discussion}
In this section, we further analyze how the quantity of synthetic data impacts performance, and what types of errors are represented in each of the MLM and NMT portions of the synthetic data.
\subsection{Amount of Synthetic Data}
As previously observed, the amount of synthetic data is orders of magnitude larger than the amount of human-annotated data. It begs the question: How much benefit do we get from smaller amounts of synthetic data? To analyze how the quantity of synthetic data affects QE performance, we conduct an ablation study of word-level QE.\footnote{The ablation study is only trained on word-level data.} As shown in Table~\ref{tab:data-size}, using only about half of the synthetic data generated (200k for En-De and 100k for En-Zh) is comparable to using the full generated set. While this suggests an upper-bound in performance to training on synthetic data. The ablation also suggests that this synthetic process can yield good performance with even a small amount of synthetic data.

\begin{table}[t]\small
    \centering
    \begin{tabular}{rr ccc}\toprule[1pt]  
        &{\bf size} & {\bf MCC} & {\bf F1-BAD} & {\bf F1-OK}\\
        \midrule   
        \multicolumn{5}{l}{\bf English-German}\\
        &100k & 37.72 & 46.23 & 83.99\\
        &200k & 38.45 & 46.79 & 84.27 \\
        
        &All (459k) & 38.68 & 46.78 & 83.85\\
        \midrule
        \multicolumn{5}{l}{\bf  English-Chinese}\\
        &50k & 53.07 & 66.58 & 83.65 \\
        &100k & 53.88 & 67.13 & 84.10\\
        &All (189k) & 53.42 & 66.86 & 84.47\\
        \bottomrule[1pt]
    \end{tabular}
    \caption{Ablation study of synthetic data amounts.}
    \label{tab:data-size}
\end{table}



\subsection{Error Analysis}
In addition to the performance, we posit that there are essential differences between MLM and NMT synthetic data. To test that, bilingual volunteers qualitatively analyzed the types of mistakes from MLM rewrites vs traditional NMT translations. The major reported differences in error types are:

\begin{enumerate}
    \item Deletions from NMT translations appear more natural and do not destroy the sentence fluency. However, deletions in MLM rewrites are more destructive (e.g., ``new york restaurants'' vs ``new restaurants'' The semantics is changed).
    \item  Most incorrect insertions or deletions from NMT translations are due to re-ordering words. (e.g., ``on 2020 in california'' vs ``in california on 2020'') However insertions with MLM-rewrites introduces seemingly random words. 
    \item  NMT translations often have semantically distant word substitutions. However, MLM-rewrites tend to substitute similar words (e.g., ``strong tea" vs ``powerful tea").
\end{enumerate}

In summary, NMT translations and MLM-rewrites appear to generate different types of errors -- the former leads to more subtle errors while the latter often introduces more catastrophic errors. Since a high-quality QE model should be able to detect both types of errors, ensembling the models trained on these two forms of synthetic data indeed is expected to outperform using only one form of synthetic data.

\section{Conclusions and Future Work}
In this work we devise a technique for building word and sentence-level QE models by creating synthetic training data. By training an off-the-shelf model on our synthetic data, we achieve performance comparable to and often better than training on human-annotated data. This technique for data synthesis can be invaluable if human annotation is difficult to come-by, for example when dealing with low-resource scenarios.

This work can be extended in various ways. While we investigate the scenario of utilizing solely synthetic data, further work can study the effects of augmenting human-labeled data with synthetic data. Further work can analyze the efficacy of this technique into low-resource language pairs where such human-annotation is difficult to obtain. Additionally, instead of a simple MLM re-writer, adversarial training to generate and detect errors could provide more realistic synthetic data.

\balance

\bibliography{eacl2021}
\bibliographystyle{acl_natbib}

\clearpage
\appendix

\end{document}